\documentclass[11pt]{article}
\pdfoutput=1
\usepackage[preprint]{acl}
\usepackage{times}
\usepackage{latexsym}
\usepackage[T1]{fontenc}
\usepackage[utf8]{inputenc}
\usepackage{microtype}
\usepackage{inconsolata}
\usepackage{amsmath}
\usepackage{amssymb}
\usepackage{amsthm}
\usepackage{graphicx}
\usepackage{booktabs}
\usepackage{multirow}
\usepackage{placeins}

\newtheorem{theorem}{Theorem}
\newtheorem{proposition}[theorem]{Proposition}
\newtheorem{corollary}[theorem]{Corollary}
\theoremstyle{definition}
\newtheorem{definition}[theorem]{Definition}

\newcommand{\eos}{\dashv}
\newcommand{\Vt}{\widetilde{\mathcal{V}}}
\newcommand{\bth}{\boldsymbol{\theta}}
\newcommand{\bq}{\mathbf{q}}
\newcommand{\bs}{\mathbf{s}}
\newcommand{\syl}{\mathrm{syl}}
\newcommand{\FKGL}{\mathrm{FKGL}}

\title{Flesch--Kincaid Readability Depends Only on the Topic Distribution\\ in Long Texts under Topic Models}

\author{Yo Ehara \\
  Tokyo Gakugei University \\
  Tokyo, Japan \\
  \texttt{ehara@u-gakugei.ac.jp}}

\begin{document}
\maketitle

\begin{abstract}
Flesch Reading Ease (FRE) and the Flesch--Kincaid Grade Level (FKGL) are computed from
the same two document statistics, and stability on long documents need not imply
invariance to lexical composition. Under a topic model with an explicit
sentence-boundary token, these scores converge almost surely to a deterministic function
of the document topic distribution; the theory covers both formulae, while the
experiments evaluate FKGL. For admixture models the limit
depends on two linear functionals; product-of-experts models share the factorisation. In
a fixed admixture with $\operatorname{rank}[\mathbf 1,\bq,\bs]=3$, fibres through
interior topic vectors are locally $(K-3)$-dimensional, whereas regular iso-score level
sets are locally $(K-2)$-dimensional and curved. Out of fold on Brown and the written
BNC, a topic vector inferred from one document half's content words predicts the other
half's FKGL at $r=0.779$ and $0.884$. On Brown, adding the topic prediction to genre and
mean content-word syllable count yields $\Delta R^2=0.002$, with a confidence interval
spanning zero; on the BNC, the corresponding split-half increment is $0.024$, positive in
four of five $K{=}100$ fits (median $0.021$). Long-text scores are thus strongly
predictable from lexical composition, with increments varying by corpus and model fit.
We do not evaluate human readability or causal effects.
\end{abstract}

\section{Introduction}
\label{sec:intro}

\begin{figure}[t]
\centering
\includegraphics[width=0.94\columnwidth]{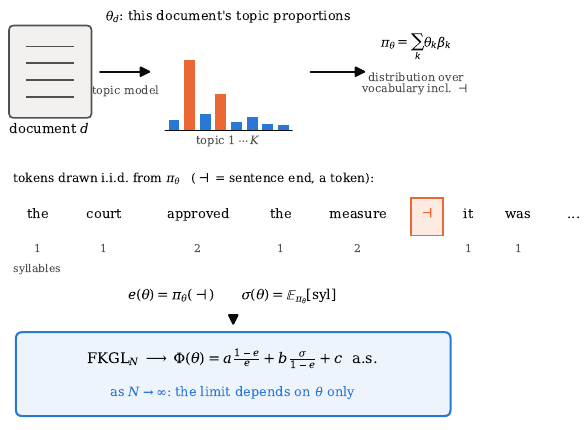}
\caption{The setting and the claim. A topic model assigns each document $d$ its own topic
proportions $\bth_d$; $\bth$ mixes per-topic word distributions into a token distribution
$\pi_{\bth}$ over the vocabulary \emph{including} a sentence-boundary token $\eos$. Both
FKGL terms are ratios of token-level sample means, so as the text grows FKGL converges
a.s.\ to a closed-form function of $\bth$ (Theorem~\ref{thm:main}).}
\label{fig:teaser}
\end{figure}

Flesch Reading Ease (FRE; \citealp{flesch1948}) and the Flesch--Kincaid Grade Level
(FKGL; \citealp{kincaid1975}), which we collectively call the Flesch--Kincaid
readability formulae, are widely used metrics for English readability
\citep{si2001,collins2005}. Both are computed from exactly the same two document
statistics: words per sentence and syllables per word. For word, sentence and syllable
counts $W$, $S$ and $Y$, their common form is
\begin{equation}
R_{a,b,c} \;\equiv\; a\,\frac{W}{S} + b\,\frac{Y}{W} + c ,
\label{eq:fkgl}
\end{equation}
with $(a,b,c)=(-1.015,-84.6,206.835)$ for FRE and $(0.39,11.8,-15.59)$ for FKGL. Given
$W/S$ and $Y/W$, either formula converts exactly into the other by replacing these three
fixed coefficients; no further document information is required (a single score alone
does not determine the other). We therefore derive the common theory once and use
FKGL notation thereafter; the formal results cover FRE after coefficient substitution,
while the corpus experiments concern FKGL.

The formulae require no word list and are computable in one pass over raw text; FKGL
additionally expresses its output as a US school grade. They remain in wide use and
increasingly serve as automatic evaluation signals for controllable generation and
simplification with large language models
\citep{tanprasert2021,imperial2023flesch,kew2023bless}. Practitioners know FKGL is
unstable on short texts, and the standard remedy is to aggregate: score a long document
and treat the stabilised value as a property of its writing. Stability with length,
however, need not imply invariance to lexical composition. We ask what the score
converges to under an explicit generative model, taking LDA \citep{blei2003lda}, ETM
\citep{dieng2020etm} or ProdLDA \citep{srivastava2017avitm} as that model.

Under a topic model with an explicit sentence-boundary token, the long-text score
converges almost surely not to a document-independent constant plus noise, but to a
deterministic function $\Phi(\bth)$ of the document topic distribution vector (the
topic vector) $\bth$, a probability vector with $\theta_k\ge0$ and
$\sum_{k}\theta_k=1$. Throughout, ``depends
only on $\bth$'' means that once the topic-model decoder, vocabulary, boundary
convention, syllable map and formula coefficients are fixed, $\bth$ is the limit's only
document-varying argument. The mechanism is elementary: a word's syllable count is a
property of its type, and words per sentence is the reciprocal of a boundary-symbol
rate, so both terms of Eq.~\eqref{eq:fkgl} are ratios of token-level sample means, which
converge conditionally on $\bth$. Beyond this law-of-large-numbers limit, we derive its
explicit two-rate factorisation, invariance geometry and asymptotic variance.

Two clarifications delimit the claim. First, a bag-of-words topic vector absorbs
whatever document-persistent variation the model can represent --- genre, register and
style along with subject matter --- so the result concerns the model variable $\bth$,
not semantics alone. Second, it does not show that either formula fails to measure
readability: lexical composition may reflect genuine difficulty, as when children's
vocabulary makes prose easier; our corpora contain no human judgements, and the
experiments identify no causal effects. The structural conclusion is that, under the
model, long-text Flesch--Kincaid scores are determined by document-level lexical
composition through only two scalar rates (Fig.~\ref{fig:teaser}).

\paragraph{Contributions.}
\begin{itemize}
\itemsep0em
\item Using FKGL notation, under any topic model with conditionally i.i.d.\ tokens,
$\FKGL_N \to \Phi(\bth)$ almost surely, in closed form for admixture and
product-of-experts models (Theorem~\ref{thm:main}, Cor.~\ref{cor:admixture},
Prop.~\ref{prop:prodlda}), with $\FKGL_N-\Phi(\bth)=O_p(N^{-1/2})$ and an explicit
asymptotic variance (Theorem~\ref{thm:rate}); the results apply to FRE after inserting
its coefficients.
\item The limit factors through two scalars: the boundary-token rate and the expected
syllable contribution per token. For a fixed $K$-topic admixture with
$\operatorname{rank}[\mathbf 1,\bq,\bs]=3$, fibres through interior topic vectors are
locally $(K-3)$-dimensional, while regular iso-score level sets are locally
$(K-2)$-dimensional and curved transverse to those fibres (Prop.~\ref{prop:geom});
along a two-topic mixing path, $\Phi$ is a ratio of polynomials of degree at most two,
not an average of the endpoint scores (Prop.~\ref{prop:mix}).
\item We audit this dependence out of fold on Brown and the written BNC under masked
inputs, disjoint document halves, genre controls and a one-feature lexical baseline. A
topic vector inferred from one half's content words predicts the other half's FKGL at
$r=0.779$ (Brown) and $0.884$ (BNC). On Brown, adding the topic prediction to genre and
mean content-word syllable count yields $\Delta R^2=0.002$, with a confidence interval
spanning zero; on the BNC the increment is $0.024$ in the default fit and positive in
four of five $K{=}100$ fits (median $0.021$): the pipeline adds no detectable value over
the baseline on Brown, while the modest BNC increment depends on the fit.
\end{itemize}

\section{Related Work}
\label{sec:related}

\paragraph{Readability formulae and their critique.}
Classical formulae combine a sentence-length term with a word-difficulty term
\citep{flesch1948,kincaid1975,smith1967ari,coleman1975}. Their limitations are well
documented: insensitivity to grammatical and discourse difficulty, gameability, and
weak correlation with human judgements
\citep{tanprasert2021,vajjala2022survey}. Modern readability assessment uses supervised
models over rich features or contextual encoders
\citep{si2001,collins2005,pitler2008,martinc2021,lee2021,crossley2023}; that formulae
transfer poorly across genres is known and usually treated as a calibration caveat
\citep{sheehan2014,xia2016}. \citet{attia2023} predict human-annotated readability from
document-level statistical and lexical features; our object is instead the FKGL formula
itself. Closest to us, \citet{belem2025} report that human readability
judgements themselves shift with topic and information content, and \citet{cachola2025}
analyse the mismatch between FKGL and human judgements in plain-language summarisation;
both study judgements or deployments; we derive the sensitivity as a structural property
of the formula under an explicit generative model.

\paragraph{Rate-based readings of FKGL.}
Both terms of Eq.~\eqref{eq:fkgl} are reciprocals of token rates, and reading them that way
is useful in itself. Treat a document as a token sequence in which
every sentence ends with an explicit boundary symbol $\eos$, and write
$p_{sw}=\#\text{sentences}/\#\text{words}$, $p_{w\ell}=\#\text{words}/\#\text{syllables}$
and $p_{s\ell}=p_{sw}p_{w\ell}$. Then Eq.~\eqref{eq:fkgl} rearranges exactly, with no
approximation, into
\begin{equation}
\FKGL - c \;=\; \frac{1}{p_{s\ell}}\bigl(a\,p_{w\ell} + b\,p_{sw}\bigr) \;=\; \frac{M}{p_{s\ell}} ,
\label{eq:ehara}
\end{equation}
with $M \equiv a p_{w\ell} + b p_{sw}$; for \emph{observed} documents every word carries
at least one syllable and every segmented sentence at least one word --- our preprocessing
appends $\eos$ only to non-empty sentences, so the counts satisfy
$0<S\le W\le Y$ --- hence $0<p_{w\ell},p_{sw}\le1$ and $|M|\le|a|+|b|$ (a property of
observed counts, not of the generative model, whose i.i.d.\ boundary process can emit
empty sentences): for either formula, the score minus its intercept is the average
number of syllables per sentence rescaled by a bounded, coefficient-dependent factor. The factorisation
in Eq.~\eqref{eq:ehara}, the boundedness of $M$, and a BNC-based analysis of these
rates were established in prior work \citep[see also \citealp{ehara2023icce}]{ehara2024paclic}.
Beyond those results, this paper introduces the token-exchangeable generative model,
the limit $\Phi(\bth)$ and its fibre and level-set geometry, the asymptotic variance,
and the masked out-of-fold audit.
Cross-genre bias of readability formulae was measured directly by \citet{sheehan2013}, and
\citet{goldstein2008} manipulated topic and genre and reported effects on Flesch scores;
we add the generative account of why they are structural.

\paragraph{Topic models.}
LDA \citep{blei2003lda} represents a document as a mixture $\pi_{\bth}=\sum_k \theta_k
\boldsymbol{\beta}_k$ over topic--word distributions. Neural topic models replace the
Dirichlet posterior with an amortised logistic-normal one
\citep{kingma2014vae,miao2016nvdm}: ProdLDA \citep{srivastava2017avitm} replaces the
mixture with a product of experts, $\pi_{\bth}=\mathrm{softmax}(\widetilde{\boldsymbol\beta}^{\top}\bth)$,
while ETM \citep{dieng2020etm} keeps the mixture but parameterises each topic by an
embedding, $\beta_{kv}\propto\exp(\boldsymbol{\rho}_v^{\top}\boldsymbol{\alpha}_k)$. All three
share the property our theory needs: a document-level simplex vector $\bth$ inducing a
\emph{probability distribution} $\pi_{\bth}$ with tokens conditionally i.i.d.\ given it,
which is what makes $e(\bth)$ and $\sigma(\bth)$ well defined. NMF can be normalised into
an admixture, truncated SVD in general cannot; both serve only as reconstruction controls
(App.~\ref{app:scope}).

\section{FKGL under Topic Models}
\label{sec:method}

\subsection{Setup}

Let $\mathcal{V}$ be a word vocabulary and $\Vt = \mathcal{V}\cup\{\eos\}$ the vocabulary
augmented with a sentence-boundary token. Let $\syl:\mathcal{V}\to[1,s_{\max}]$ be a
bounded syllable count --- a deterministic property of the word type --- extended by
$\syl(\eos)=0$. (Non-integer values are allowed because the experiments assign an averaged
count to an \texttt{<unk>} type; only boundedness and $\syl\ge1$ on words are used.) All
probabilistic statements below are conditional on $\bth$.

\begin{definition}[Token-exchangeable topic model]
\label{def:model}
A \emph{token-exchangeable topic model} on $\Vt$ consists of a document-level variable
$\bth \in \Delta^{K-1}$ with some prior, and a map $\bth \mapsto \pi_{\bth} \in
\Delta^{|\Vt|-1}$, such that the tokens $v_1,\dots,v_N$ of a document are i.i.d.\ draws
from $\pi_{\bth}$.
\end{definition}

LDA and ETM are of this form with the \emph{admixture} map
$\pi_{\bth}(v)=\sum_{k=1}^{K}\theta_k \beta_{kv}$; ProdLDA is of this form with the
\emph{product-of-experts} map $\pi_{\bth}=\mathrm{softmax}(\widetilde{\boldsymbol\beta}^{\top}\bth + \mathbf{b})$.

Letting the same document-level mixture emit sentence boundaries is a modelling
assumption, and a consequential one: the count of $\eos$ tokens is, together with the
syllable-weighted word counts, a sufficient statistic for Eq.~\eqref{eq:fkgl}. We return to
what this costs us in Sec.~\ref{sec:whattheta} and test it empirically in
Sec.~\ref{sec:exp-main}. We define, for any $\bth$, the two \emph{token functionals}
\begin{align}
e(\bth) &\equiv \pi_{\bth}(\eos) , \label{eq:e}\\
\sigma(\bth) &\equiv \sum_{w\in\mathcal{V}} \pi_{\bth}(w)\,\syl(w) . \label{eq:sigma}
\end{align}
$e(\bth)$ is the probability that a token is a sentence boundary; $\sigma(\bth)$ is the
expected number of syllables contributed by a token.

\subsection{The long-text limit}

\begin{theorem}[FKGL is a function of $\bth$ alone]
\label{thm:main}
Let a document of $N$ tokens be generated by a token-exchangeable topic model with
$0<e(\bth)<1$. Let $\FKGL_N$ denote Eq.~\eqref{eq:fkgl} computed on that document. Then,
almost surely as $N\to\infty$,
\begin{align}
\FKGL_N &\;\longrightarrow\; \Phi(\bth) , \notag \\
\Phi(\bth) &\;\equiv\; a\,\frac{1-e(\bth)}{e(\bth)} \;+\; b\,\frac{\sigma(\bth)}{1-e(\bth)} \;+\; c .
\label{eq:main}
\end{align}
\end{theorem}

The proof (App.~\ref{app:proofs}) is the strong law of large numbers applied to the
bounded i.i.d.\ sentence, word and syllable counts, followed by the continuous mapping
theorem.

For finite $N$, $S_N=0$ or $W_N=0$ ($\FKGL_N$ undefined) occurs with probability
$(1-e)^N+e^N\to0$; assigning any fixed value on this vanishing event leaves the limit
unchanged (the simulation of App.~\ref{app:rate} discards such replicates, which arise
only at $N\le50$).

\begin{proposition}[Dependent tokens]
\label{prop:ergodic}
Suppose that, conditionally on $\bth$, the token process $(v_n)_{n\ge1}$ is stationary and
ergodic with one-dimensional marginal $\pi_{\bth}$. Then the conclusion of
Theorem~\ref{thm:main} holds, with the same $\Phi$.
\end{proposition}

The proof is identical: only two sample averages need to converge, and Birkhoff's theorem
supplies that. This covers burstiness, $n$-gram dependence and sentence-length
autocorrelation. Within-document topic drift is a special case with a twist. Conditional
on the drift $(\bth_n)$, let $v_n\sim\pi_{\bth_n}$ independently (bounded martingale
differences suffice) and let the occupation measures
$\mu_N=N^{-1}\sum_{n\le N}\delta_{\bth_n}\Rightarrow\mu$ weakly with
$\bar e\equiv\int e\,d\mu\in(0,1)$. For
admixture models $\pi_{\bth}$ is linear in $\bth$, so the document converges to
$\Phi(\bar\bth)$ with $\bar\bth=\int\bth\,d\mu$; for ProdLDA the
softmax is non-linear, and the averaged rate pair $(\bar e,\bar\sigma)$ need not lie in
the image of any single topic vector.

\subsection{Closed form for admixture models}

\begin{corollary}[LDA, ETM]
\label{cor:admixture}
For an admixture model $\pi_{\bth}=\sum_k \theta_k \boldsymbol{\beta}_k$, define the
per-topic scalars
\begin{equation}
q_k \equiv \beta_{k\eos}, \qquad s_k \equiv \sum_{w\in\mathcal{V}}\beta_{kw}\syl(w) ,
\label{eq:qs}
\end{equation}
and collect them into $\bq,\bs\in\mathbb{R}^{K}$. Then $e(\bth)=\bq^{\top}\bth$ and
$\sigma(\bth)=\bs^{\top}\bth$; writing
$\mathcal{D}\equiv\{\bth\in\Delta^{K-1}:0<\bq^{\top}\bth<1\}$ for the set on which $\Phi$
is defined, for every $\bth\in\mathcal{D}$
\begin{equation}
\Phi(\bth) = a\,\frac{1-\bq^{\top}\bth}{\bq^{\top}\bth} + b\,\frac{\bs^{\top}\bth}{1-\bq^{\top}\bth} + c .
\label{eq:admix}
\end{equation}
\end{corollary}

In the notation of Eq.~\eqref{eq:ehara}, the average number of syllables per sentence
becomes a ratio of two linear forms, $1/p_{s\ell}(\bth)=\bs^{\top}\bth/\bq^{\top}\bth$,
and $\Phi(\bth)-c=M(\bth)\,\bs^{\top}\bth/\bq^{\top}\bth$: Eq.~\eqref{eq:ehara} with
every quantity determined by the topic vector. For ETM the same expressions hold with
$\beta_{kv}=\exp(\boldsymbol{\rho}_v^{\top}\boldsymbol\alpha_k)/Z_k$ substituted into
Eq.~\eqref{eq:qs}.

\begin{proposition}[ProdLDA]
\label{prop:prodlda}
For $\pi_{\bth}=\mathrm{softmax}(\widetilde{\boldsymbol\beta}^{\top}\bth+\mathbf{b})$, write
$\ell_v(\bth) = (\widetilde{\boldsymbol\beta}^{\top}\bth)_v + b_v$. Then
\begin{align}
e(\bth) &= \frac{\exp \ell_{\eos}(\bth)}{\sum_{v\in\Vt}\exp \ell_v(\bth)}, \notag\\
\sigma(\bth) &= \frac{\sum_{w\in\mathcal{V}} \syl(w) \exp \ell_w(\bth)}{\sum_{v\in\Vt}\exp \ell_v(\bth)},
\end{align}
and $\Phi$ is given by Eq.~\eqref{eq:main}. The outer form is unchanged, but $e$ and
$\sigma$ are no longer linear in $\bth$, so Eq.~\eqref{eq:admix} does not apply.
\end{proposition}

\subsection{What the limit discards}
\label{sec:geom}

\begin{corollary}[Two-scalar factorisation]
\label{cor:bottleneck}
$\Phi = g \circ (e,\sigma)$ with $g(u,z) = a(1-u)/u + bz/(1-u) + c$. Hence
$\Phi(\bth)=\Phi(\bth')$ whenever $e(\bth)=e(\bth')$ and $\sigma(\bth)=\sigma(\bth')$.
\end{corollary}

Corollary~\ref{cor:bottleneck} says that FKGL reads a $K$-coordinate document
description and keeps two numbers; ``$K-2$ dimensions discarded'' conflates two
different objects, and the correct statement is sharper.

\begin{proposition}[Invariance geometry, admixture case]
\label{prop:geom}
Let $\mathbf{M}=[\mathbf{1},\bq,\bs]^{\top}\in\mathbb{R}^{3\times K}$ with
$\rho=\operatorname{rank}\mathbf{M}$, and let
$\mathcal{F}_\Delta(u,z)=\{\bth\in\Delta^{K-1}: \bq^{\top}\bth=u,\ \bs^{\top}\bth=z\}$ be
the set of \emph{valid} topic vectors sharing the two rates.
\begin{enumerate}
\itemsep0em
\item[(i)] For $u\in(0,1)$ with $\mathcal{F}_\Delta(u,z)\neq\emptyset$:
$\mathcal{F}_\Delta(u,z)$ is a convex polytope on which $\Phi$ is constant; its
affine hull has dimension at most $K-\rho$, and when $(u,z)$ is attained at a point of
$\operatorname{relint}\Delta^{K-1}$ the polytope has that full local dimension --- for
$\rho=3$, exactly $K-3$.
\item[(ii)] At any $\bth\in\operatorname{relint}\Delta^{K-1}\cap\mathcal{D}$ with
$g_e\bq+g_\sigma\bs \notin \operatorname{span}\{\mathbf{1}\}$, the level set of $\Phi$ in
the simplex is locally a smooth manifold of dimension $K-2$, with tangent space
$\{\boldsymbol\delta: \mathbf{1}^{\top}\boldsymbol\delta=0,\
(g_e\bq+g_\sigma\bs)^{\top}\boldsymbol\delta=0\}$.
\item[(iii)] If $\rho=3$: for any convex set
$C\subset\mathcal{D}$ on which $\Phi$ is constant, the image of
$C$ under $\bth\mapsto(\bq^{\top}\bth,\bs^{\top}\bth)$ is a convex subset of a strictly
concave level curve of $g$, hence a singleton; so $C\subset\mathcal{F}_\Delta(u,z)$ for a
single $(u,z)$. Level sets therefore contain no line segment transverse to the fibres:
they are curved.
\end{enumerate}
\end{proposition}

Parts (i) and (ii) are immediate; part (iii) follows because the level curves of $g$ are
strictly concave and so contain no line segment (App.~\ref{app:proofs}); ``regular''
throughout means condition (ii), regularity of the simplex-restricted differential. The
two notions
come apart: the valid reallocations FKGL discards by construction form a polytope of
local dimension generically $K-3$, while the FKGL-preserving perturbations at a given
document form one dimension more, $K-2$, whose extra direction rotates with $\bth$.
These are properties of the parameterisation, not
invariants of the generative distribution --- duplicating a topic raises $K$ and the fibre
dimension without changing $\pi_{\bth}$ --- hence the rank and interiority qualifiers. For ProdLDA
the analogue of (ii) holds locally, with $\nabla\Phi=g_e\nabla e+g_\sigma\nabla\sigma$,
wherever $\nabla\Phi(\bth)\notin\operatorname{span}\{\mathbf{1}\}$. Fibres need not be
convex polytopes, but a regular $(e,\sigma)$-fibre is still locally a $(K-3)$-manifold
wherever the simplex-restricted Jacobian of $(e,\sigma)$ has rank two (constant-rank
theorem); what need not carry over is the convexity of (i) and the no-transverse-segment
conclusion of (iii), since $e,\sigma$ are not linear in $\bth$.

\subsection{Mixing two topics}

\begin{proposition}[Two-topic mixing path]
\label{prop:mix}
For an admixture model and $\bth_\lambda = (1-\lambda)\mathbf{e}_j + \lambda \mathbf{e}_k$,
write $q_\lambda=(1-\lambda)q_j+\lambda q_k$ and $s_\lambda=(1-\lambda)s_j+\lambda s_k$,
and suppose $q_\lambda\in(0,1)$ for all $\lambda\in[0,1]$ (e.g.\ $q_j,q_k\in(0,1)$).
Then
\begin{equation}
\Phi(\bth_\lambda) - c \;=\; \frac{a(1-q_\lambda)^{2} + b\,s_\lambda q_\lambda}{q_\lambda(1-q_\lambda)} ,
\label{eq:mix}
\end{equation}
a ratio of polynomials of degree at most two in $\lambda$. In particular
$\Phi(\bth_\lambda)$ is in
general neither affine nor linear-fractional in $\lambda$, and
$\Phi(\bth_\lambda)$ need not equal $(1-\lambda)\Phi(\mathbf{e}_j)+\lambda\Phi(\mathbf{e}_k)$.
\end{proposition}

Each term of Eq.~\eqref{eq:mix} is separately linear-fractional in $\lambda$, but their
sum has type $(2,2)$, not $(1,1)$, so $\Phi$ along the path is \emph{not} a M\"obius
transform; only the syllables-per-sentence factor
$1/p_{s\ell}(\bth_\lambda)=s_\lambda/q_\lambda$ is. Practically: a chapter half easy
narrative and half hard exposition does not receive the average grade.

\begin{corollary}[When is the statement vacuous?]
\label{cor:vacuous}
For an admixture model with $q_k\in(0,1)$ for all $k$ --- so that $\Phi$ is defined on
all of $\Delta^{K-1}$ --- $\Phi$ is constant on $\Delta^{K-1}$ if and only if
$q_k \equiv q$ and $s_k \equiv s$ for all $k$.
\end{corollary}

The proof (App.~\ref{app:proofs}) maps the simplex to
$\operatorname{conv}\{(q_k,s_k)\}_k$, which constancy of $\Phi$ confines to a strictly
concave level curve, hence to a point. Corollary~\ref{cor:vacuous} characterises the restrictive condition under which FKGL
would be invariant to topic mixture in this fixed model: $(q_k,s_k)$ constant across
topics. Sec.~\ref{sec:exp-topics} measures the spread.

\subsection{Rate of convergence}

\begin{theorem}[Asymptotic normality]
\label{thm:rate}
Under the conditions of Theorem~\ref{thm:main}, conditionally on $\bth$ and with
$\sigma_2(\bth)\equiv\sum_{w}\pi_{\bth}(w)\syl(w)^2$,
\begin{equation}
\sqrt{N}\bigl(\FKGL_N - \Phi(\bth)\bigr) \;\Rightarrow\; \mathcal{N}\!\left(0,\;\tau^2(\bth)\right),
\end{equation}
\begin{align}
\tau^2 &= g_e^2\,e(1-e) \;-\; 2\,g_e g_\sigma\, e\,\sigma \;+\; g_\sigma^2(\sigma_2-\sigma^2), \notag\\
g_e &= -\frac{a}{e^{2}} + \frac{b\,\sigma}{(1-e)^{2}}, \qquad g_\sigma = \frac{b}{1-e} .
\label{eq:tau}
\end{align}
In particular $\FKGL_N-\Phi(\bth)=O_p(N^{-1/2})$.
\end{theorem}

The proof (App.~\ref{app:proofs}) applies the multivariate CLT and the delta method to the
bounded i.i.d.\ vector $T_n=(\mathbf{1}[v_n=\eos],\syl(v_n))$, whose covariance is
determined by $e,\sigma,\sigma_2$ because $\syl(\eos)=0$. (On the undefined event
$\FKGL_N$ is fixed arbitrarily, as after Theorem~\ref{thm:main}.) The error is $O_p$, not
almost-sure, and its constant blows up as $e\to0$: texts with long sentences need more
tokens for FKGL to stabilise, because the sentence-count term is the reciprocal of a
rare-event rate.

\subsection{What $\bth$ is, and what it is not}
\label{sec:whattheta}

Three qualifications precede any interpretation. \emph{First}, $\bth$ is not semantics: a
bag-of-words topic model has exactly one document-level latent, so everything persistent is
pushed into it. A model with a semantic $\bth$ and a separate style variable $\mathbf{z}$
would give $\Phi(\bth,\mathbf{z})$ by the same proof, and nothing says the $\mathbf{z}$
dependence is weak. \emph{Second}, putting $\eos$ in the vocabulary is not innocuous. The
$\eos$ count \emph{is} the sentence count, so an inference network that sees the augmented
bag of words has been handed one of the two sufficient statistics of Eq.~\eqref{eq:fkgl}
directly; no readability \emph{supervision} enters the model, but that is much weaker than
no readability \emph{information}. Sec.~\ref{sec:exp-main} therefore re-estimates $\bth$
with $\eos$, and then all NLTK stopword-list types, masked from the inference input.
\emph{Third}, the measurable quantity is not the theorem: a corpus experiment can only ask
how much of the measured FKGL is recoverable from a fitted $K$-dimensional
representation, controlled against non-topic-model representations
(Sec.~\ref{sec:exp-main}).

\section{Experiments}
\label{sec:exp}

Theorem~\ref{thm:main} is an asymptotic statement about an idealised generative process,
and no corpus experiment can verify it directly. What an experiment can measure is how much
of the FKGL of a held-out document is recoverable from a topic vector inferred by a model
fitted without it --- and how much of that is recoverable from far
cruder summaries of the same lexical information. We therefore evaluate against masked
inference inputs, across disjoint document halves, under genre control, and against a
one-dimensional lexical baseline, jointly rather than one at a time.

\subsection{Setup}
\label{sec:setup}

\paragraph{Corpora and preprocessing.} We use two corpora. The \textbf{Brown} corpus
\citep{francis1979brown}, via NLTK \citep{bird2009nltk}: 500 documents of about 2{,}000
words (1{,}005{,}119 word tokens) in 15 categories grouped into informative and imaginative
prose. The written \textbf{BNC} \citep{bnc2007}: every written text with $\ge50$ sentences,
giving 3{,}021 documents, 86.7M word tokens, median 31{,}040 words per document ---
fifteen times Brown. Genre controls pool Lee's \citeyearpar{lee2001} super-genres
with $n{<}20$ as ``other''; within-genre summaries use the 12 with $n{\ge}20$.
On both corpora we keep every
token containing a letter, case-fold only, remove no stopwords, and append an explicit
$\eos$ per sentence; types below a frequency floor (5 / 10) map to \texttt{<unk>} with the
frequency-weighted mean syllable count of the absorbed types (Brown $2.663$, $5.55\%$ of
tokens; BNC $2.715$, $1.46\%$); $|\Vt|=13{,}976$ and $75{,}002$. Syllables come from
CMUdict \citep[version 0.7a, via NLTK;][]{cmudict} with a vowel-group fallback. Computing
FKGL from raw surface forms instead changes nothing (App.~\ref{app:ci}). The vocabulary, frequency floor and \texttt{<unk>} syllable
value are computed corpus-wide, so ``out of fold'' below refers to model fitting, not to
preprocessing.

\paragraph{Models.} On Brown: \textbf{LDA} \citep{blei2003lda}, \textbf{ProdLDA}
\citep{srivastava2017avitm} and \textbf{ETM} \citep{dieng2020etm}, $K=50$. On the BNC: LDA
with $K=100$, within the range of the original studies (up to 100 in \citealp{blei2003lda};
50--200 in \citealp{srivastava2017avitm}; 50--300 in \citealp{dieng2020etm}) --- a
capacity choice for the larger corpus, not claimed optimal (fit sensitivity in
Sec.~\ref{sec:exp-bc} and App.~\ref{app:k}); hyperparameters in App.~\ref{app:hp}. Each fitted model yields
$\hat\bth_d$ and $\hat\pi_d=\pi_{\hat\bth_d}$, from which $\Phi$ is read off
(Cor.~\ref{cor:admixture}, Prop.~\ref{prop:prodlda}); matrix factorisations serve only as
reconstruction baselines (App.~\ref{app:scope}).

\paragraph{Protocol.} All headline predictive results (Tables~\ref{tab:main},
\ref{tab:dr2}, and App.~\ref{app:ci}--\ref{app:decomp}) are out of fold: 5-fold
cross-validation over a seed-0 random partition (not genre-stratified), the model ---
hence $(\bq,\bs)$ or the decoder --- fitted only on the training folds; explicitly
labelled topic illustrations and $K$/seed sensitivity analyses use whole-corpus fits.
``Not seen'' means the decoder was fitted without the document; $\hat\bth$ is always
inferred from (part of) its own counts. Half A is the
contiguous first half of each document, cut at the median sentence boundary, half B the
remainder; the same fixed partition serves all models and only the A$\to$B direction is
evaluated (chosen ex ante). The genre-mean baseline is fitted on outer training folds
only. Genre-residual correlations use category means estimated on the outer training
folds (out-of-fold centring); the whole-sample descriptive variant is shown alongside.
Document-level bootstrap 95\% intervals (2{,}000 resamples) are conditional on the fitted
folds, seed and decoder (they do not propagate fold-partition or fitting variation).
Correlations and RMSEs in
Table~\ref{tab:main} are uncalibrated; the regression combinations of
Sec.~\ref{sec:exp-bc} are cross-fitted.

\paragraph{Inference-input variants and baselines.} Because the $\eos$ count \emph{is} the
sentence count (Sec.~\ref{sec:whattheta}), we use three inference inputs: \emph{full};
\emph{no}\,$\eos$ (the query document's $\eos$ coordinate zeroed); and \emph{content}
($\eos$ and all 198 NLTK stopwords zeroed). Masking removes the boundary and stopword
\emph{coordinates} (198 NLTK stopword-list
types, not all function words), while content-word identities still encode syllable
length: the content variant tests whether lexical composition suffices to infer the
omitted rates, not whether semantics determines FKGL. Masking is test-time only, but a
word-only model reproduces the result (Sec.~\ref{sec:exp-bc}); either way the regularity
is empirical, not a consequence of Theorem~\ref{thm:main}. To calibrate how much of the
signal is trivially lexical, we also evaluate
$B_C$, the mean syllable count of a text's content words: a single number, computable
without any model.

\subsection{Are topics FKGL-neutral?}
\label{sec:exp-topics}

Corollary~\ref{cor:vacuous} says Theorem~\ref{thm:main} has content only to the extent
that $(q_k,s_k)$ vary across topics. They vary: in the $K=50$ Brown LDA model, the 13
topics with $q_k\ge0.01$ carry $97.8\%$ of posterior topic mass, spanning implied
sentence lengths of $9.6$--$46.7$ words and $1.24$--$1.80$ syllables per word
(Table~\ref{tab:topics}: dialogue and narration at the low end;
government, technical and mathematical prose at the high end). Cautions: pure-topic
vertices are unoccupied (observed $\Phi(\hat\bth_d)$ spans $4.5$--$14.1$ at the 1st--99th
percentile for this fit; App.~\ref{app:robust}) and per-topic values are only moderately
seed-stable (App.~\ref{app:robust}).

\subsection{Out-of-fold agreement}
\label{sec:exp-main}

\begin{table}[t]
\centering\small
\setlength{\tabcolsep}{2.1pt}
\begin{tabular}{@{}lcccccc@{}}
\toprule
& \multicolumn{3}{c}{whole document} & \multicolumn{3}{c}{half A $\to$ B} \\
\cmidrule(lr){2-4}\cmidrule(l){5-7}
model & full & no\,$\eos$ & cont. & full & no\,$\eos$ & cont. \\
\midrule
\multicolumn{7}{@{}l}{\emph{Brown} (500 docs, $\sim$2k words, $K{=}50$)} \\
LDA & \textbf{0.861} & 0.853 & \textbf{0.848} & 0.786 & 0.779 & 0.779 \\
ProdLDA & 0.809 & 0.797 & 0.797 & 0.751 & 0.739 & 0.745 \\
ETM & 0.805 & 0.790 & 0.765 & 0.743 & 0.730 & 0.700 \\
$B_C$ (1 feature) & 0.867 & --- & --- & 0.784 & --- & --- \\
measured half A & --- & --- & --- & 0.880 & --- & --- \\
category mean & 0.736 & --- & --- & --- & --- & --- \\
\midrule
\multicolumn{7}{@{}l}{\emph{written BNC} (3{,}021 docs, median 31k words, $K{=}100$)} \\
LDA & \textbf{0.928} & 0.920 & \textbf{0.908} & 0.900 & 0.892 & 0.884 \\
$B_C$ (1 feature) & 0.888 & --- & --- & 0.845 & --- & --- \\
measured half A & --- & --- & --- & 0.943 & --- & --- \\
genre mean & 0.813 & --- & --- & --- & --- & --- \\
\bottomrule
\end{tabular}
\caption{Out-of-fold Pearson $r$ between predictions and measured FKGL (5-fold; the decoder
is always fitted without the evaluated document). \emph{full}: inference input is the whole
$\eos$-augmented bag of words; \emph{no}\,$\eos$: its $\eos$ coordinate zeroed;
\emph{cont.}: $\eos$ and all 198 NLTK stopwords zeroed. $B_C$ = mean content-word syllable
count of the input text (of half A in the split columns). CIs, slopes, RMSE and $R^2$ for
the LDA rows in App.~\ref{app:ci}.}
\label{tab:main}
\end{table}

Out of fold, LDA correlates with measured FKGL at $r=0.861$ (CI $[0.833,0.887]$) on Brown
and $0.928$ (CI $[0.922,0.933]$) on the BNC; RMSEs $1.70$ and $1.27$ grades. The
prediction plots further show that the fitted values are less dispersed than the
measured FKGL scores, especially near the extremes (App.~\ref{app:ci},
Fig.~\ref{fig:scatter}). Masking barely moves the topic models: $\eos$ zeroed,
$0.861\to0.853$ and $0.928\to0.920$; stopword types zeroed too, $0.848$ and $0.908$.
Predicting the FKGL \emph{measured on half B} from $\hat\bth$ inferred on half A gives
$0.786/0.779$ (Brown, full/content) and $0.900/0.884$ (BNC), against test--retest
reference correlations of $0.880$ and $0.943$ (a different protocol). The strongest
single number: \textbf{a topic vector inferred from one half's content words predicts
the other half's measured FKGL at $r=0.884$} (BNC).

\subsection{How much of this is trivially lexical?}
\label{sec:exp-bc}

\begin{table}[t]
\centering\small
\setlength{\tabcolsep}{2.6pt}
\begin{tabular}{@{}lccc@{}}
\toprule
& \multicolumn{2}{c}{Brown} & BNC \\
\cmidrule(lr){2-3}\cmidrule(l){4-4}
features (from half A) & FKGL$_B$ & (W/S)$_B$ & FKGL$_B$ \\
\midrule
genre dummies                  & 0.511 & 0.307 & 0.651 \\
genre $+\,B_C$                 & 0.660 & 0.335 & 0.785 \\
genre $+\,B_C+\Phi(\hat\bth)$  & 0.661 & 0.337 & 0.810 \\
$\Phi(\hat\bth)$ alone         & 0.599 & ---   & 0.781 \\
\bottomrule
\end{tabular}
\caption{Cross-fitted out-of-fold $R^2$ of linear models predicting half-B FKGL (Brown and
BNC) and half-B words-per-sentence (Brown) from half-A features; $\Phi(\hat\bth)$ is the
content-masked topic prediction. Split-half $\Delta R^2$ of adding $\Phi$ to
genre$+B_C$: $0.002$ $[-0.003,0.007]$ on Brown, $0.024$ $[0.018,0.031]$ on the BNC. In
the less stringent whole-document analyses (full-input prediction) the increments are
$0.010$ $[0.000,0.020]$ and $0.041$ $[0.035,0.048]$. BNC fit sensitivity in
App.~\ref{app:decomp}.}
\label{tab:dr2}
\end{table}

The $B_C$ rows of Table~\ref{tab:main} are the sobering calibration: on Brown, $B_C$
alone predicts whole-document FKGL at $0.867$ and half-B FKGL at $0.784$, matching
the topic pipeline ($0.861$, $0.779$). Cross-fitted models sharpen this
(Table~\ref{tab:dr2}): genre
dummies give out-of-fold $R^2=0.51$, genre $+\,B_C$ $0.66$, and adding $\Phi(\hat\bth)$
$0.66$; paired document bootstrap puts the $\Phi$--$B_C$ correlation difference at $-0.007$
$[-0.034,0.020]$ and the $\Delta R^2$ at $0.002$ $[-0.003,0.007]$, so ``matches'' is
statistical. Brown restarts $0/1/2$ on fixed folds give $0.779$--$0.785$.
Component-level evaluation (Table~\ref{tab:comp}) locates the signal: from content-masked
half A, half B's syllable rate is recovered at $r=0.833$, near its test--retest reference
correlation of $0.887$; its sentence-length rate only at $0.552$ (reference correlation
$0.812$): on Brown,
what the representation carries is essentially the lexical component. A word-only LDA
with a rate-supervised ridge probe $\bth\to(e,\sigma)$ (App.~\ref{app:decomp}) reaches
$r=0.845$ on the content-masked whole-document task, against $0.848$ boundary-augmented
($0.911$ vs $0.908$ on the BNC) --- though the probe uses measured training-document
rates as supervision, so it is not equivalent to the unsupervised decoder pipeline. On
Brown, then, the topic pipeline recovers FKGL no better than average content-word length
plus genre --- an empirical regularity, not a consequence of Theorem~\ref{thm:main}. The
BNC differs: $\Phi$ beats $B_C$ by $\Delta r=0.039$ $[0.031,0.048]$, and under the
content-masked split-half protocol adding $\Phi$ to genre and $B_C$ yields
$\Delta R^2=0.024$ $[0.018,0.031]$ ($0.041$ $[0.035,0.048]$ in the less stringent
whole-document, full-input analysis; Tables~\ref{tab:dr2}, \ref{tab:bnc}). All headline
BNC numbers use the default fit ($K{=}100$, seed 0), fixed in advance. Because
variational LDA reaches restart-dependent local optima, we refitted four further
$K{=}100$ fits and one $K{=}50$ fit (App.~\ref{app:decomp}): the whole-document
increment is positive in all six fits ($0.038$--$0.074$), and the split-half increment
is positive in the $K{=}50$ fit and four of the five $K{=}100$ fits (median $0.021$),
reversing for one boundary-dominated fit; the intervals above are conditional on the
fitted model (Sec.~\ref{sec:setup}). Unlike Brown, the BNC shows a positive incremental
signal in four of the five $K{=}100$ fits, although its magnitude and even its sign are
not fully robust to model fitting.

\subsection{Is it just genre?}
\label{sec:exp-genre}

With out-of-fold centring, the genre-residual correlation on whole documents, full
input, is $0.670$ on Brown (whole-sample variant $0.656$) and $0.768$ on the BNC
($0.771$); with content masking and split-half applied \emph{simultaneously} it is
$0.475$ ($0.460$) on Brown --- against $0.549$ for $B_C$ and $0.740$ for measured half
A --- and $0.638$ ($0.643$) on the BNC (App.~\ref{app:decomp}). BNC within-genre
correlations have a Fisher-$z$, size-weighted mean of $0.784$ over the 12 super-genres
with $n\ge20$ (range $0.520$--$0.912$); genre-mean baselines: $0.736$/$0.813$. These are separate summaries, not a decomposition:
strong between genres, moderate within; on Brown the within-genre part is carried at
least as well by $B_C$.

\subsection{Text length}
\label{sec:exp-length}

Agreement rises with segment length on both corpora (Fig.~\ref{fig:length},
App.~\ref{app:length}): on Brown
from $0.556$ (25 words) to $0.858$ (1{,}600); on the BNC from $0.587$ to $0.930$
(12{,}800), with Spearman $\rho$ monotone throughout. This is descriptive: the theorem
concerns one document's convergence, not cross-document correlation; sampling error of
$\FKGL_N$ and estimation error of $\hat\bth$ shrink together.

\subsection{Convergence rate and plug-in bias}
\label{sec:exp-rate}

\begin{figure}[t]
\centering
\includegraphics[width=0.65\columnwidth]{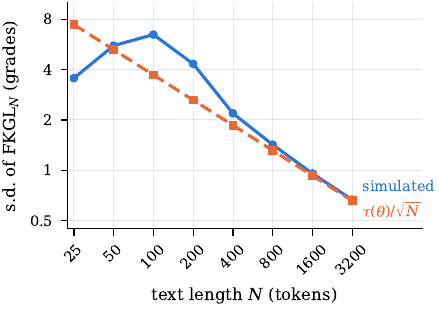}
\caption{Finite-sample validation of the asymptotic s.d.\ of Theorem~\ref{thm:rate}:
observed s.d.\ of $\FKGL_N$ under i.i.d.\ resampling against $\tau(\bth)/\sqrt{N}$ (60
Brown documents, 300 replicates; App.~\ref{app:rate}). Within ${\sim}3\%$ of the
empirical s.d.\ at $N=1{,}600$; underestimates variability for $N\le200$.}
\label{fig:rate}
\end{figure}

Fig.~\ref{fig:rate} shows the asymptotic standard deviation becoming accurate for long
texts: within ${\sim}3\%$ of the empirical s.d.\ at $N=1{,}600$, but substantially
underestimating variability for $N\le200$. We therefore read the closed-form variance as
a long-text approximation, not a finite-sample guarantee. Posterior-predictive averaging over 64 draws changes
correlations by $\le0.002$ (App.~\ref{app:rate}); this does not bound bias in grade
units, correlation being insensitive to additive and scale bias.

\section{Discussion}
\label{sec:disc}

\paragraph{What the experiments license.}
FKGL is strongly predictable, out of fold, from the document-level lexical distribution
a topic model compresses; the association is not exhausted by coarse genre labels. On
Brown, the incremental signal is consistently largely exhausted by genre plus one
lexical scalar ($B_C$); on the BNC, this conclusion depends on the fitted topic model,
with a positive incremental signal in four of the five $K{=}100$ fits.
This does not say FKGL fails to measure readability --- nor that it is useless: genre
centring leaves a substantial within-genre association with lexical composition, whose
topical and stylistic components we do not separate.

\paragraph{Consequences.}
FKGL is routinely used to check that a language model writes at a requested level,
increasingly as an automatic reward (Sec.~\ref{sec:intro}).
A generator can lower FKGL by shifting its lexical distribution towards smaller $\Phi$,
or move along fibre directions with FKGL constant (Prop.~\ref{prop:geom}); whether such
shifts are simplification FKGL cannot say, likewise for FKGL-band filtering.

\section{Conclusion}
\label{sec:conc}

Under a topic model with conditionally i.i.d.\ tokens, FKGL converges a.s.\ to a
closed-form function of the topic vector with $O_p(N^{-1/2})$ error; for
$\operatorname{rank}[\mathbf 1,\bq,\bs]=3$, interior fibres are locally $(K-3)$- and
regular level sets $(K-2)$-dimensional and curved. A topic vector from
one half's content words predicts the other half's FKGL at $r=0.884$ (BNC) and $0.779$
(Brown); a one-feature baseline matches this on Brown; the BNC increment is
fit-dependent, positive in four of five $K{=}100$ fits.

\section*{Limitations}

\paragraph{Theory.} The conditional i.i.d.\ assumption is false of real text.
Prop.~\ref{prop:ergodic} relaxes it to stationarity and ergodicity given $\bth$ (almost-sure
convergence only; the $O_p(N^{-1/2})$ rate and Theorem~\ref{thm:rate} are proved under
independence and will understate the variance under sentence-length autocorrelation). The
drift extension requires convergence of the empirical occupation measures; without it,
$\FKGL_N$ need not converge. The
geometry of Prop.~\ref{prop:geom} is a property of a fixed admixture parameterisation, not
of the generative distribution.

\paragraph{$\bth$ is not semantics.} A bag-of-words topic model has one document-level
latent, so genre, register, style and typographic convention are absorbed into $\bth$ with
subject matter, and Sec.~\ref{sec:exp-bc} shows that on Brown the recoverable signal is
matched by mean content-word syllable length plus genre. Nothing in our data separates
subject matter from style; a model with separate latents would yield
$\Phi(\bth,\mathbf{z})$ by the same proof.

\paragraph{The $\eos$ construction.} Placing $\eos$ in the vocabulary is what defines the
first term of Eq.~\eqref{eq:fkgl} under a bag-of-words model, and it hands the model one of
the two sufficient statistics of FKGL; the masked variants change the inference input only,
and $q_k$ is still estimated from training documents in which $\eos$ was observed.

\paragraph{Empirical scope.} Two corpora, one language; neither is controlled for topic ---
graded readers or paired simplification corpora remain the natural next experiment. The
BNC audit is limited to LDA, and variational LDA reaches a restart-dependent local
optimum: across five $K{=}100$ fits and one $K{=}50$ fit, the whole-document increment
is positive in all six, while the content-masked split-half increment is positive in the
$K{=}50$ fit and four of the five $K{=}100$ fits but negative in the remaining $K{=}100$
fit (App.~\ref{app:decomp}); bootstrap intervals are conditional on the fitted model,
and fold-partition sensitivity remains untested. Vocabulary and \texttt{<unk>} statistics are
corpus-wide, so preprocessing is not fold-separated. Segment bootstrap intervals in
Fig.~\ref{fig:length} ignore document clustering. Brown LDA at $K=50$ effectively used 13
topics, and per-topic quantities are only moderately stable across seeds
(App.~\ref{app:robust}). $\Phi(\hat\bth)$ is a diagnostic, not a proposed readability
predictor.

\section*{Ethical Considerations}

Our analysis is mathematical and our experiments use two long-standing, licensed corpora of
published English, Brown and the British National Corpus; we see no direct ethical risk in
the work itself. One downstream risk deserves note: FKGL responds to document-level lexical
composition, so using it as an automatic reward or filter can systematically disadvantage
texts with particular lexical profiles --- which correlate with subject matter and genre
--- regardless of qualities our data cannot measure. Practitioners who gate content on FKGL
should be aware of what the score actually tracks. Regarding release: the BNC licence does
not permit redistribution of the corpus text, so we will release code, document identifiers
and derived statistics only, sufficient to reproduce every number from a licensed copy.

\section*{Acknowledgements}

This work was supported by JSPS KAKENHI Grant Numbers JP22K12287 and JP26K15085,
and by JST, PRESTO Grant Number JPMJPR2363.

\bibliography{custom}

\begin{thebibliography}{41}
\providecommand{\natexlab}[1]{#1}

\bibitem[{Attia et~al.(2023)Attia, Samih, and Ehara}]{attia2023}
Mohammed Attia, Younes Samih, and Yo~Ehara. 2023.
\newblock Statistical measures for readability assessment.
\newblock In \emph{Proceedings of the Joint 3rd International Conference on
  Natural Language Processing for Digital Humanities and 8th International
  Workshop on Computational Linguistics for Uralic Languages}, pages 153--161,
  Tokyo, Japan. Association for Computational Linguistics.

\bibitem[{Belem et~al.(2025)Belem, Glenn, Samuel, Kumar, and Liu}]{belem2025}
Catarina Belem, Parker Glenn, Alfy Samuel, Anoop Kumar, and Daben Liu. 2025.
\newblock Readability reconsidered: A cross-dataset analysis of reference-free
  metrics.
\newblock In \emph{Proceedings of the Fourth Workshop on Text Simplification,
  Accessibility and Readability (TSAR 2025)}, pages 47--69, Suzhou, China.
  Association for Computational Linguistics.

\bibitem[{Bird et~al.(2009)Bird, Klein, and Loper}]{bird2009nltk}
Steven Bird, Ewan Klein, and Edward Loper. 2009.
\newblock \emph{Natural Language Processing with {Python}}.
\newblock O'Reilly Media.

\bibitem[{Blei et~al.(2003)Blei, Ng, and Jordan}]{blei2003lda}
David~M. Blei, Andrew~Y. Ng, and Michael~I. Jordan. 2003.
\newblock Latent {Dirichlet} allocation.
\newblock \emph{Journal of Machine Learning Research}, 3:993--1022.

\bibitem[{{BNC Consortium}(2007)}]{bnc2007}
{BNC Consortium}. 2007.
\newblock The {British} {National} {Corpus}, {XML} edition.
\newblock Oxford Text Archive.

\bibitem[{Cachola et~al.(2025)Cachola, Khashabi, and Dredze}]{cachola2025}
Isabel Cachola, Daniel Khashabi, and Mark Dredze. 2025.
\newblock Evaluating the evaluators: Are readability metrics good measures of
  readability?
\newblock In \emph{Proceedings of the 2025 Conference on Empirical Methods in
  Natural Language Processing}, pages 24011--24027, Suzhou, China. Association
  for Computational Linguistics.

\bibitem[{Coleman and Liau(1975)}]{coleman1975}
Meri Coleman and T.~L. Liau. 1975.
\newblock \href {https://doi.org/10.1037/h0076540} {A computer readability
  formula designed for machine scoring}.
\newblock \emph{Journal of Applied Psychology}, 60(2):283--284.

\bibitem[{Collins-Thompson and Callan(2005)}]{collins2005}
Kevyn Collins-Thompson and Jamie Callan. 2005.
\newblock \href {https://doi.org/10.1002/asi.20243} {Predicting reading
  difficulty with statistical language models}.
\newblock \emph{Journal of the American Society for Information Science and
  Technology}, 56(13):1448--1462.

\bibitem[{Crossley et~al.(2023)Crossley, Heintz, Choi, Batchelor, Karimi, and
  Malatinszky}]{crossley2023}
Scott Crossley, Aron Heintz, Joon~Suh Choi, Jordan Batchelor, Mehrnoush Karimi,
  and Agnes Malatinszky. 2023.
\newblock A large-scaled corpus for assessing text readability.
\newblock \emph{Behavior Research Methods}, 55(2):491--507.

\bibitem[{Dieng et~al.(2020)Dieng, Ruiz, and Blei}]{dieng2020etm}
Adji~B. Dieng, Francisco J.~R. Ruiz, and David~M. Blei. 2020.
\newblock Topic modeling in embedding spaces.
\newblock \emph{Transactions of the Association for Computational Linguistics},
  8:439--453.

\bibitem[{Ehara(2023)}]{ehara2023icce}
Yo~Ehara. 2023.
\newblock \href {https://doi.org/10.58459/icce.2023.1480} {A novel
  interpretation of classical readability metrics: Revisiting the language
  model underpinning the {Flesch-Kincaid} index}.
\newblock In \emph{Proceedings of the 31st International Conference on
  Computers in Education (Work-in-Progress Poster)}, pages 939--941.
  Asia-Pacific Society for Computers in Education.

\bibitem[{Ehara(2024)}]{ehara2024paclic}
Yo~Ehara. 2024.
\newblock An analytical study of the {Flesch-Kincaid} readability formulae to
  explain their robustness over time.
\newblock In \emph{Proceedings of the 38th Pacific Asia Conference on Language,
  Information and Computation}, pages 989--997, Tokyo, Japan. Tokyo University
  of Foreign Studies.

\bibitem[{Flesch(1948)}]{flesch1948}
Rudolf Flesch. 1948.
\newblock \href {https://doi.org/10.1037/h0057532} {A new readability
  yardstick}.
\newblock \emph{Journal of Applied Psychology}, 32(3):221--233.

\bibitem[{Francis and Ku{\v{c}}era(1979)}]{francis1979brown}
W.~Nelson Francis and Henry Ku{\v{c}}era. 1979.
\newblock \emph{Manual of Information to Accompany a Standard Corpus of
  Present-Day Edited {American} {English}, for use with Digital Computers}.
\newblock Department of Linguistics, Brown University.

\bibitem[{Goldstein-Stewart et~al.(2008)Goldstein-Stewart, Goodwin, Sabin, and
  Winder}]{goldstein2008}
Jade Goldstein-Stewart, Kerri Goodwin, Roberta Sabin, and Ransom Winder. 2008.
\newblock Creating and using a correlated corpus to glean communicative
  commonalities.
\newblock In \emph{Proceedings of the Sixth International Conference on
  Language Resources and Evaluation (LREC)}.
\newblock \url{https://aclanthology.org/L08-1198/}.

\bibitem[{Harris et~al.(2020)Harris, Millman, van~der Walt, Gommers, Virtanen,
  Cournapeau, Wieser, Taylor, Berg, Smith, Kern, Picus, Hoyer, van Kerkwijk,
  Brett, Haldane, del R{\'i}o, Wiebe, Peterson, G{\'e}rard-Marchant, Sheppard,
  Reddy, Weckesser, Abbasi, Gohlke, and Oliphant}]{harris2020array}
Charles~R. Harris, K.~Jarrod Millman, St{\'e}fan~J. van~der Walt, Ralf Gommers,
  Pauli Virtanen, David Cournapeau, Eric Wieser, Julian Taylor, Sebastian Berg,
  Nathaniel~J. Smith, Robert Kern, Matti Picus, Stephan Hoyer, Marten~H. van
  Kerkwijk, Matthew Brett, Allan Haldane, Jaime~Fern{\'a}ndez del R{\'i}o, Mark
  Wiebe, Pearu Peterson, and 7 others. 2020.
\newblock \href {https://doi.org/10.1038/s41586-020-2649-2} {Array programming
  with {NumPy}}.
\newblock \emph{Nature}, 585(7825):357--362.

\bibitem[{Hoffman et~al.(2010)Hoffman, Bach, and Blei}]{hoffman2010olda}
Matthew~D. Hoffman, Francis~R. Bach, and David~M. Blei. 2010.
\newblock Online learning for latent {Dirichlet} allocation.
\newblock \emph{Advances in Neural Information Processing Systems},
  23:856--864.

\bibitem[{Imperial and Tayyar~Madabushi(2023)}]{imperial2023flesch}
Joseph~Marvin Imperial and Harish Tayyar~Madabushi. 2023.
\newblock Flesch or fumble? {E}valuating readability standard alignment of
  instruction-tuned language models.
\newblock In \emph{Proceedings of the Third Workshop on Natural Language
  Generation, Evaluation, and Metrics (GEM)}, pages 205--223.

\bibitem[{Kew et~al.(2023)Kew, Chi, V{\'a}squez-Rodr{\'\i}guez, Agrawal,
  Aumiller, Alva-Manchego, and Shardlow}]{kew2023bless}
Tannon Kew, Alison Chi, Laura V{\'a}squez-Rodr{\'\i}guez, Sweta Agrawal, Dennis
  Aumiller, Fernando Alva-Manchego, and Matthew Shardlow. 2023.
\newblock {BLESS}: Benchmarking large language models on sentence
  simplification.
\newblock In \emph{Proceedings of the 2023 Conference on Empirical Methods in
  Natural Language Processing}, pages 13291--13309.

\bibitem[{Kincaid et~al.(1975)Kincaid, Fishburne, Rogers, and
  Chissom}]{kincaid1975}
J.~Peter Kincaid, Robert~P. Fishburne, Jr., Richard~L. Rogers, and Brad~S.
  Chissom. 1975.
\newblock \href {https://doi.org/10.21236/ADA006655} {Derivation of new
  readability formulas (automated readability index, fog count and flesch
  reading ease formula) for {Navy} enlisted personnel}.
\newblock Research Branch Report 8-75, Naval Technical Training Command,
  Millington TN Research Branch.

\bibitem[{Kingma and Welling(2014)}]{kingma2014vae}
Diederik~P. Kingma and Max Welling. 2014.
\newblock Auto-encoding variational {Bayes}.
\newblock In \emph{International Conference on Learning Representations}.

\bibitem[{Lee et~al.(2021)Lee, Jang, and Lee}]{lee2021}
Bruce~W. Lee, Yoo~Sung Jang, and Jason Hyung-Jong Lee. 2021.
\newblock Pushing on text readability assessment: A transformer meets
  handcrafted linguistic features.
\newblock In \emph{Proceedings of the 2021 Conference on Empirical Methods in
  Natural Language Processing}, pages 10669--10686.

\bibitem[{Lee(2001)}]{lee2001}
David Y.~W. Lee. 2001.
\newblock Genres, registers, text types, domains and styles: Clarifying the
  concepts and navigating a path through the {BNC} jungle.
\newblock \emph{Language Learning \& Technology}, 5(3):37--72.

\bibitem[{Martinc et~al.(2021)Martinc, Pollak, and
  Robnik-{\v{S}}ikonja}]{martinc2021}
Matej Martinc, Senja Pollak, and Marko Robnik-{\v{S}}ikonja. 2021.
\newblock Supervised and unsupervised neural approaches to text readability.
\newblock \emph{Computational Linguistics}, 47(1):141--179.

\bibitem[{Miao et~al.(2016)Miao, Yu, and Blunsom}]{miao2016nvdm}
Yishu Miao, Lei Yu, and Phil Blunsom. 2016.
\newblock Neural variational inference for text processing.
\newblock In \emph{Proceedings of the 33rd International Conference on Machine
  Learning}, pages 1727--1736.

\bibitem[{Paszke et~al.(2019)Paszke, Gross, Massa, Lerer, Bradbury, Chanan,
  Killeen, Lin, Gimelshein, Antiga, Desmaison, K{\"o}pf, Yang, DeVito, Raison,
  Tejani, Chilamkurthy, Steiner, Fang, Bai, and
  Chintala}]{NEURIPS2019_bdbca288}
Adam Paszke, Sam Gross, Francisco Massa, Adam Lerer, James Bradbury, Gregory
  Chanan, Trevor Killeen, Zeming Lin, Natalia Gimelshein, Luca Antiga, Alban
  Desmaison, Andreas K{\"o}pf, Edward Yang, Zachary DeVito, Martin Raison,
  Alykhan Tejani, Sasank Chilamkurthy, Benoit Steiner, Lu~Fang, and 2 others.
  2019.
\newblock \href
  {https://proceedings.neurips.cc/paper_files/paper/2019/file/bdbca288fee7f92f2bfa9f7012727740-Paper.pdf}
  {{PyTorch}: An imperative style, high-performance deep learning library}.
\newblock In \emph{Advances in Neural Information Processing Systems},
  volume~32. Curran Associates, Inc.

\bibitem[{Pedregosa et~al.(2011)}]{pedregosa2011sklearn}
Fabian Pedregosa et~al. 2011.
\newblock Scikit-learn: Machine learning in {Python}.
\newblock \emph{Journal of Machine Learning Research}, 12:2825--2830.

\bibitem[{Pitler and Nenkova(2008)}]{pitler2008}
Emily Pitler and Ani Nenkova. 2008.
\newblock Revisiting readability: A unified framework for predicting text
  quality.
\newblock In \emph{Proceedings of the 2008 Conference on Empirical Methods in
  Natural Language Processing}, pages 186--195.

\bibitem[{{\v R}eh{\r u}{\v r}ek and Sojka(2010)}]{rehurek2010gensim}
Radim {\v R}eh{\r u}{\v r}ek and Petr Sojka. 2010.
\newblock Software framework for topic modelling with large corpora.
\newblock In \emph{Proceedings of the LREC 2010 Workshop on New Challenges for
  NLP Frameworks}, pages 45--50.

\bibitem[{Sheehan et~al.(2013)Sheehan, Flor, and Napolitano}]{sheehan2013}
Kathleen~M. Sheehan, Michael Flor, and Diane Napolitano. 2013.
\newblock A two-stage approach for generating unbiased estimates of text
  complexity.
\newblock In \emph{Proceedings of the Workshop on Natural Language Processing
  for Improving Textual Accessibility}, pages 49--58, Atlanta, Georgia.
  Association for Computational Linguistics.
\newblock \url{https://aclanthology.org/W13-1506/}.

\bibitem[{Sheehan et~al.(2014)Sheehan, Kostin, Napolitano, and
  Flor}]{sheehan2014}
Kathleen~M. Sheehan, Irene Kostin, Diane Napolitano, and Michael Flor. 2014.
\newblock The {TextEvaluator} tool: Helping teachers and test developers select
  texts for use in instruction and assessment.
\newblock \emph{The Elementary School Journal}, 115(2):184--209.

\bibitem[{Si and Callan(2001)}]{si2001}
Luo Si and Jamie Callan. 2001.
\newblock \href {https://doi.org/10.1145/502585.502695} {A statistical model
  for scientific readability}.
\newblock In \emph{Proceedings of the Tenth International Conference on
  Information and Knowledge Management}, pages 574--576.

\bibitem[{Sievert and Shirley(2014)}]{sievert2014}
Carson Sievert and Kenneth Shirley. 2014.
\newblock {LDAvis}: A method for visualizing and interpreting topics.
\newblock In \emph{Proceedings of the Workshop on Interactive Language
  Learning, Visualization, and Interfaces}, pages 63--70, Baltimore, Maryland,
  USA. Association for Computational Linguistics.

\bibitem[{Smith and Senter(1967)}]{smith1967ari}
Edgar~A. Smith and R.~J. Senter. 1967.
\newblock Automated readability index.
\newblock Technical Report AMRL-TR-66-220, Aerospace Medical Research
  Laboratories.

\bibitem[{Srivastava and Sutton(2017)}]{srivastava2017avitm}
Akash Srivastava and Charles Sutton. 2017.
\newblock Autoencoding variational inference for topic models.
\newblock In \emph{International Conference on Learning Representations}.

\bibitem[{Tanprasert and Kauchak(2021)}]{tanprasert2021}
Teerapaun Tanprasert and David Kauchak. 2021.
\newblock {Flesch-Kincaid} is not a text simplification evaluation metric.
\newblock In \emph{Proceedings of the 1st Workshop on Natural Language
  Generation, Evaluation, and Metrics (GEM 2021)}, pages 1--14.

\bibitem[{{The joblib developers}(2025)}]{joblib2025}
{The joblib developers}. 2025.
\newblock \href {https://doi.org/10.5281/zenodo.17936197} {{joblib}}.
\newblock Zenodo.
\newblock Version 1.5.3.

\bibitem[{Vajjala(2022)}]{vajjala2022survey}
Sowmya Vajjala. 2022.
\newblock Trends, limitations and open challenges in automatic readability
  assessment research.
\newblock In \emph{Proceedings of the Thirteenth Language Resources and
  Evaluation Conference (LREC)}, pages 5366--5377.

\bibitem[{Virtanen et~al.(2020)Virtanen, Gommers, Oliphant, Haberland, Reddy,
  Cournapeau, Burovski, Peterson, Weckesser, Bright, {van der Walt}, Brett,
  Wilson, Millman, Mayorov, Nelson, Jones, Kern, Larson, Carey, Polat, Feng,
  Moore, {VanderPlas}, Laxalde, Perktold, Cimrman, Henriksen, Quintero, Harris,
  Archibald, Ribeiro, Pedregosa, {van Mulbregt}, and {SciPy 1.0
  Contributors}}]{2020SciPy-NMeth}
Pauli Virtanen, Ralf Gommers, Travis~E. Oliphant, Matt Haberland, Tyler Reddy,
  David Cournapeau, Evgeni Burovski, Pearu Peterson, Warren Weckesser, Jonathan
  Bright, St{\'e}fan~J. {van der Walt}, Matthew Brett, Joshua Wilson, K.~Jarrod
  Millman, Nikolay Mayorov, Andrew R.~J. Nelson, Eric Jones, Robert Kern, Eric
  Larson, and 16 others. 2020.
\newblock \href {https://doi.org/10.1038/s41592-019-0686-2} {{SciPy} 1.0:
  Fundamental algorithms for scientific computing in python}.
\newblock \emph{Nature Methods}, 17:261--272.

\bibitem[{Weide(2008)}]{cmudict}
Robert~L. Weide. 2008.
\newblock The {Carnegie Mellon University} pronouncing dictionary, version
  0.7a.
\newblock Carnegie Mellon University.
\newblock \url{http://www.speech.cs.cmu.edu/cgi-bin/cmudict}.

\bibitem[{Xia et~al.(2016)Xia, Kochmar, and Briscoe}]{xia2016}
Menglin Xia, Ekaterina Kochmar, and Ted Briscoe. 2016.
\newblock Text readability assessment for second language learners.
\newblock In \emph{Proceedings of the 11th Workshop on Innovative Use of NLP
  for Building Educational Applications}, pages 12--22.

\end{thebibliography}

\appendix

\section{Hyperparameters}
\label{app:hp}

All models use $K=50$ topics unless stated otherwise and are trained on the same
$\eos$-augmented count matrix of 500 documents over $|\Vt|=13{,}976$ types.

\paragraph{LDA.} Batch variational Bayes as implemented in scikit-learn
\citep{pedregosa2011sklearn} following \citet{hoffman2010olda}; symmetric document--topic
and topic--word priors both set to $1/K$; 200 EM iterations. $\hat\bth$ is the normalised
variational Dirichlet mean returned by \texttt{transform}.

\paragraph{ProdLDA and ETM.} Both use the same inference network: two fully connected
layers of width 400 with softplus activations, dropout $0.1$ (ProdLDA: $0.2$), and linear
heads producing the mean and log-variance of a logistic-normal posterior, each followed by
a non-affine batch-normalisation layer. $\bth=\mathrm{softmax}(z)$ with $z$ reparameterised.
Training uses Adam ($\beta_1=0.99$) with batch size 128. ProdLDA is trained for 300 epochs
at learning rate $2\times10^{-3}$ with the KL term annealed linearly over the first 100
epochs; its decoder is $\mathrm{softmax}(\widetilde{\boldsymbol\beta}^{\top}\bth+\mathbf{b})$
with $\mathbf{b}$ initialised to the log unigram distribution (we use this per-type bias in
place of the usual decoder batch-normalisation, which distorts the absolute token
probabilities that Eq.~\eqref{eq:main} requires). ETM is trained for 600 epochs at learning
rate $5\times10^{-3}$ with KL annealing over 200 epochs; its word embeddings
$\boldsymbol{\rho}\in\mathbb{R}^{|\Vt|\times128}$ are initialised from skip-gram vectors
trained on the same $\eos$-augmented corpus (window 5, 15 epochs, \texttt{gensim}
\citep{rehurek2010gensim}) and fine-tuned. At evaluation time $\hat\bth=\mathrm{softmax}(\mu)$.
The skip-gram initialisation is fitted on the whole corpus, so ETM's preprocessing is
transductive; the decoder and inference network themselves are fitted per fold.

\paragraph{BNC.} The written BNC (XML edition) is parsed keeping texts with $\ge50$
sentences; tokens are the \texttt{<w>} elements of \texttt{<s>} units, case-folded, with
$\eos$ appended per sentence. LDA uses $K=100$, symmetric $1/K$ priors, batch variational
Bayes, 150 iterations; folds, masks, split-half and bootstrap are identical to Brown.
Genre analysis uses David Lee's classification \citep{lee2001} collapsed to super-genres;
genre controls pool those with $n<20$ documents as ``other'', and within-genre summaries
use the 12 with $n\ge20$.

\paragraph{Reproducibility.} Brown experiments run on CPU in under two hours; the BNC
experiment in under five hours on 50 cores. Seeds are fixed (0 for models, 1 for segment
subsampling). Upon acceptance, we will release code, document identifiers and derived
statistics (\texttt{results\_lda\_K100\_seed0.json} and companions);
corpus text is not redistributed (see Ethical Considerations). Artifact licences,
compute and package versions are detailed in App.~\ref{app:checklist}.
\FloatBarrier

\section{Deferred proofs}
\label{app:proofs}

\paragraph{Thm.~\ref{thm:main}.}
Write $S_N=\sum_{n=1}^{N}\mathbf{1}[v_n=\eos]$ for the number of sentences,
$W_N = N-S_N$ for the number of words, and $Y_N=\sum_{n=1}^{N}\syl(v_n)$ for the number of
syllables (the convention $\syl(\eos)=0$ makes the boundary tokens contribute nothing).
The summands are bounded i.i.d.\ random variables, so by the strong law of large numbers
$S_N/N \to e(\bth)$, $W_N/N \to 1-e(\bth)$ and $Y_N/N\to\sigma(\bth)$ almost surely,
hence $W_N/S_N \to (1-e(\bth))/e(\bth)$ and $Y_N/W_N \to \sigma(\bth)/(1-e(\bth))$
almost surely by the continuous mapping theorem. Substituting into Eq.~\eqref{eq:fkgl}
gives Eq.~\eqref{eq:main}. \qed

\paragraph{Cor.~\ref{cor:vacuous}.}
($\Leftarrow$) is immediate. For ($\Rightarrow$): the image of $\Delta^{K-1}$ under
$\bth\mapsto(\bq^{\top}\bth,\bs^{\top}\bth)$ is $\operatorname{conv}\{(q_k,s_k)\}_{k=1}^K$.
If $\Phi$ is constant, this convex set lies in one level curve of $g$, which is strictly
concave (see the proof of Prop.~\ref{prop:geom} below) and so contains no non-degenerate
segment; hence the hull is a single point and $(q_k,s_k)$ is the same for all $k$. \qed

\paragraph{Prop.~\ref{prop:geom}.} (i) and (ii) follow from the definitions and the implicit
function theorem, using $\nabla\Phi = g_e\bq+g_\sigma\bs$ with $g_e,g_\sigma$ as in
Eq.~\eqref{eq:tau}. For (iii), let $C\subseteq\mathcal{D}$ be
convex with $\Phi$ constant at value $v$ on $C$. The image of $C$
under the linear map $\bth\mapsto(\bq^{\top}\bth,\bs^{\top}\bth)$ is a convex subset of
the level curve
$\{g=v\}$, which is the graph of $h(u)=(1-u)(v-c)/b-a(1-u)^{2}/(bu)$ with
$h''(u)=-2a/(b u^{3})<0$ on $(0,1)$, since $a/b>0$ for both standard formulae (FKGL:
$a,b>0$; FRE: $a,b<0$). A strictly concave graph contains no non-degenerate
line segment, so a convex subset of it is a single point, i.e.\ $C$ lies in one fibre.

\paragraph{Thm.~\ref{thm:rate}.} The vector $T_n=(\mathbf{1}[v_n=\eos],\ \syl(v_n))$ is
i.i.d.\ and bounded with mean $(e,\sigma)$. Its covariance has entries
$\mathrm{Var}(\mathbf{1}[v_n=\eos])=e(1-e)$, $\mathrm{Var}(\syl(v_n))=\sigma_2-\sigma^2$ and
$\mathrm{Cov}=\mathbb{E}[\mathbf{1}[v_n=\eos]\syl(v_n)]-e\sigma=-e\sigma$, using
$\syl(\eos)=0$. Since $\FKGL_N=g(\bar T_N)$ and $g$ is continuously differentiable at
$(e,\sigma)$ for $0<e<1$, the multivariate CLT and the delta method give the result with
$\nabla g=(g_e,g_\sigma)$.
\FloatBarrier

\section{The two-topic mixing curve}
\label{app:mix}

\begin{figure}[h]
\centering
\includegraphics[width=0.9\columnwidth]{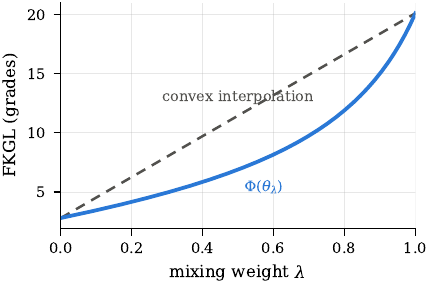}
\caption{Mixing the lowest- and highest-$\Phi$ LDA topics of Table~\ref{tab:topics}. By
Prop.~\ref{prop:mix} the curve is a ratio of two quadratics in $\lambda$; it is neither the
convex interpolation (dashed) nor a linear-fractional function.}
\label{fig:mix}
\end{figure}

Fig.~\ref{fig:mix} illustrates Prop.~\ref{prop:mix} with fitted topics.
\FloatBarrier

\section{Convergence rate and plug-in bias}
\label{app:rate}

To test Eq.~\eqref{eq:tau} free of topic-model estimation error, we take 60 random Brown
documents, treat each one's own token distribution as $\pi_{\bth}$, draw 300 i.i.d.\ texts of
$N$ tokens from it, and compare the observed standard deviation of $\FKGL_N$ with
$\tau(\bth)/\sqrt{N}$. The observed-to-predicted ratio is $1.18$ at $N=400$, $1.08$ at $800$,
$1.03$ at $1{,}600$ and $1.015$ at $3{,}200$ (Fig.~\ref{fig:rate}). Below $N\approx200$ the
delta method underestimates substantially, because the sentence count is then small and
$W_N/S_N$ is heavy-tailed --- the familiar instability of FKGL on short texts, in
quantitative form; a $0.5$-grade standard deviation needs $\sim$$5{,}700$ tokens ($\tau$
averaged over the 60 documents).

For the plug-in check we draw 64 samples from each document's posterior (the variational
Dirichlet for LDA, the logistic-normal for the neural models), evaluate $\Phi$ at each and
average. Out-of-fold correlations move from $0.861/0.809/0.805$ (plug-in) to
$0.861/0.807/0.805$ (posterior predictive) for LDA/ProdLDA/ETM.
\FloatBarrier

\section{Matrix-factorisation baselines}
\label{app:scope}

An earlier version of this work excluded NMF and truncated SVD on the grounds that they
define no probability distribution. For NMF that argument is wrong, and we retract it:
given $X\approx WH$ with non-negative factors, setting $r_k=\sum_v H_{kv}$,
$\beta_{kv}=H_{kv}/r_k$ and $\theta_{dk}=W_{dk}r_k/\sum_j W_{dj}r_j$ renormalises the
reconstruction exactly into an admixture $\sum_k\theta_{dk}\beta_{kv}$ (components with
$r_k=0$ are dropped, and each document is assumed to have positive row sum
$\sum_j W_{dj}r_j>0$), so $(q_k,s_k)$ are perfectly well defined for NMF. SVD components can be negative and admit no such
renormalisation. We therefore draw the line differently: we reserve ``generative topic
model'' for models fitted with an explicit token likelihood, and report NMF and SVD as
\emph{algebraic reconstruction baselines}.

As baselines they are informative about circularity rather than about topics. On Brown,
rank-$50$ reconstructions read FKGL off the very counts that define it: SVD reaches
$r=0.945$ and NMF $0.918$ on the unablated task, above every topic model; with the $\eos$
coordinate and stopword types deleted from the input they fall to $0.112$ and $0.023$
(topic models: $0.848$/$0.797$/$0.765$), and a random rank-$50$ subspace gives $r=-0.079$
throughout. The unablated comparison measures reconstruction fidelity; only the masked
comparison says anything about what a document-level representation carries.

\paragraph{Protocol.} Both are fitted on the outer training folds only, exactly like the
topic models, and evaluated on the held-out fold. NMF (scikit-learn, \texttt{nndsvd}
initialisation, at most $400$ iterations, tolerance $10^{-4}$) is fitted to training rows
$L^1$-normalised and rescaled to $1{,}000$ tokens; a held-out row is normalised the same
way and encoded by \texttt{transform}, i.e.\ non-negative least squares against the fixed
dictionary, and the reconstruction is $W H$. Truncated SVD is fitted to $L^1$-normalised
training rows after centring at the training mean; a held-out row is normalised, centred,
projected onto the $K$ components and reconstructed. The random baseline projects onto a
fixed rank-$50$ orthonormal subspace drawn once. In every case the (possibly negative)
reconstruction is clipped below at $10^{-12}$ and renormalised to a distribution before
$(e,\sigma)$ are read off --- an algebraic device, not a probability model, which is how a
negative SVD reconstruction yields rates at all; the comparison with the topic models'
out-of-fold numbers should be read with that in mind. Masked variants zero the specified
coordinates of the held-out row before encoding, as for the topic models.
\FloatBarrier

\section{Confidence intervals, error and calibration}
\label{app:ci}

\begin{figure}[h]
\centering
\includegraphics[width=0.86\columnwidth]{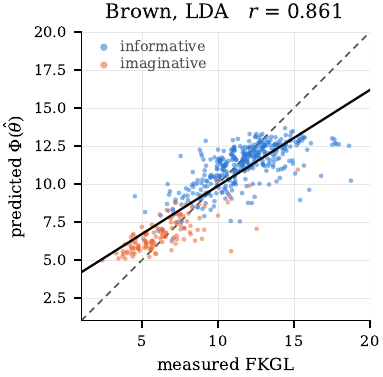}
\caption{Out-of-fold topic-predicted FKGL against measured FKGL (Brown, LDA). Dashed:
identity; solid: least-squares fit. ProdLDA/ETM look alike; numbers in
Table~\ref{tab:main}.}
\label{fig:scatter}
\end{figure}

\begin{table}[h]
\centering\small
\setlength{\tabcolsep}{1.2pt}
\begin{tabular}{@{}llcccccc@{}}
\toprule
setting & corpus & $r$ & 95\% CI & slope & RMSE & MAE & $R^2$ \\
\midrule
full, whole    & Brown & .861 & [.833,\,.887] & 1.17 & 1.70 & 1.25 & .72 \\
cont., whole   & Brown & .848 & [.817,\,.874] & 1.30 & 1.87 & 1.47 & .66 \\
full, A$\to$B  & Brown & .786 & [.746,\,.823] & 1.10 & 2.10 & 1.57 & .61 \\
cont., A$\to$B & Brown & .779 & [.742,\,.814] & 1.21 & 2.22 & 1.73 & .57 \\
\midrule
full, whole    & BNC & .928 & [.922,\,.933] & 1.08 & 1.27 & 0.94 & .86 \\
cont., whole   & BNC & .908 & [.901,\,.914] & 1.09 & 1.43 & 1.06 & .82 \\
full, A$\to$B  & BNC & .900 & [.893,\,.907] & 1.08 & 1.52 & 1.14 & .81 \\
cont., A$\to$B & BNC & .884 & [.876,\,.892] & 1.09 & 1.64 & 1.25 & .77 \\
\bottomrule
\end{tabular}
\caption{LDA, out of fold, uncalibrated. Slope regresses measured on predicted; RMSE in
grade levels; $R^2=1-\mathrm{MSE}/\mathrm{Var}$. The neural Brown models are
under-dispersed (slopes $1.79$, $1.25$), the familiar shrinkage of amortised inference.
Replacing the vocabulary-consistent target by FKGL from raw surface forms (which correlate
at $r=0.9978$ on Brown) changes the Brown correlations to $0.866/0.811/0.810$
(LDA/ProdLDA/ETM), i.e.\ nothing material.}
\label{tab:ci}
\end{table}
\FloatBarrier

\section{Example topics}
\label{app:topics}

\begin{table}[h]
\centering\small
\setlength{\tabcolsep}{3.2pt}
\begin{tabular}{@{}rrrl@{}}
\toprule
$\Phi(\mathbf{e}_k)$ & w/s & syl/w & distinctive words \\
\midrule
$\phantom{0}2.80$ & $\phantom{0}9.6$ & 1.242 & him said did don't would what \\
$\phantom{0}5.36$ & $12.9$ & 1.350 & down back front head feet into \\
$\phantom{0}7.34$ & $12.1$ & 1.543 & your feed per inches pool \\
$\phantom{0}7.78$ & $16.8$ & 1.424 & af t n polynomial operator v \\
\midrule
$13.71$ & $21.9$ & 1.760 & state states government federal united \\
$14.33$ & $28.1$ & 1.606 & clay mold pieces lid design glaze \\
$19.21$ & $34.8$ & 1.799 & index stations electronic radiation \\
$20.12$ & $46.7$ & 1.483 & **zg q tangent curve c vertex \\
\bottomrule
\end{tabular}
\caption{The four lowest- and four highest-$\Phi$ Brown LDA topics ($K=50$, whole corpus),
among the 13 with $q_k\ge0.01$. $\Phi(\mathbf{e}_k)$ is the FKGL of a text generated by
topic $k$ alone --- a simplex vertex, not an observed document. Words ranked by LDAvis
relevance \citep[$\lambda=0.4$;][]{sievert2014}, function words removed for display.}
\label{tab:topics}
\end{table}
\FloatBarrier

\section{Combined controls and lexical baseline}
\label{app:decomp}

\begin{table}[h]
\centering\small
\setlength{\tabcolsep}{2.6pt}
\begin{tabular}{@{}llccc@{}}
\toprule
predictor & target & raw $r$ & g-res.\ OOF & g-res.\ \\
\midrule
$\Phi(\hat\bth)$, full        & whole   & 0.861 & 0.670 & 0.656 \\
$\Phi(\hat\bth)$, content     & whole   & 0.848 & 0.635 & 0.621 \\
$B_C$ (doc)                   & whole   & 0.867 & 0.714 & 0.705 \\
\midrule
$\Phi(\hat\bth)$, full, A     & half B  & 0.786 & 0.492 & 0.475 \\
$\Phi(\hat\bth)$, content, A  & half B  & 0.779 & 0.475 & 0.460 \\
$B_C$ (half A)                & half B  & 0.784 & 0.563 & 0.549 \\
measured FKGL of half A       & half B  & 0.880 & 0.746 & 0.740 \\
\bottomrule
\end{tabular}
\caption{Brown, LDA $K{=}50$, all out of fold. \emph{g-res.\ OOF}: Pearson $r$ after
removing category means estimated on the outer training folds from predictor and target;
\emph{g-res.}: the descriptive variant with whole-sample category means. $B_C$ is the mean
syllable count of the input text's content words.}
\label{tab:decomp}
\end{table}

\begin{table}[h]
\centering\small
\setlength{\tabcolsep}{2.6pt}
\begin{tabular}{@{}llcccc@{}}
\toprule
model & target (half B) & $r$ & g-res.\ & RMSE & $R^2$ \\
\midrule
boundary-aug. & $W/S$    & 0.552 & 0.202 & 4.63 & 0.30 \\
boundary-aug. & $Y/W$    & 0.833 & 0.603 & 0.097 & 0.56 \\
boundary-aug. & FKGL     & 0.779 & 0.460 & 2.22 & 0.57 \\
\midrule
measured half A & $W/S$  & 0.812 & 0.718 & 3.36 & 0.64 \\
measured half A & $Y/W$  & 0.887 & 0.763 & 0.070 & 0.77 \\
measured half A & FKGL   & 0.880 & 0.740 & 1.64 & 0.76 \\
\midrule
\multicolumn{6}{@{}l}{\emph{written BNC} (LDA $K{=}100$)} \\
boundary-aug. & $W/S$    & 0.762 & 0.465 & 3.88 & 0.55 \\
boundary-aug. & $Y/W$    & 0.920 & 0.803 & 0.076 & 0.72 \\
boundary-aug. & FKGL     & 0.884 & 0.643 & 1.64 & 0.77 \\
measured half A & $W/S$  & 0.900 & --- & 2.52 & 0.81 \\
measured half A & $Y/W$  & 0.942 & --- & 0.049 & 0.89 \\
measured half A & FKGL   & 0.943 & 0.839 & 1.15 & 0.89 \\
\bottomrule
\end{tabular}
\caption{Direct two-component evaluation: predicted
$\widehat{W/S}=(1-\hat e)/\hat e$ and $\widehat{Y/W}=\hat\sigma/(1-\hat e)$ from
content-masked half A against half B's measured rates (\emph{boundary-aug.}: the standard
$\eos$-augmented LDA), with the measured half-A rates as test--retest reference rows
(whole-sample genre centring, matching the g-res.\ column). On
Brown the syllable rate is recovered near its reference correlation ($0.833$ vs.\
$0.887$) but the
sentence-length rate is not ($0.552$ vs.\ $0.812$); on the BNC the sentence-length rate
recovers much further ($0.762$ vs.\ $0.900$). MAE and further detail in
\texttt{results\_lda\_K100\_seed0.json} (released upon acceptance).}
\label{tab:comp}
\end{table}

\begin{table}[h]
\centering\small
\setlength{\tabcolsep}{2.6pt}
\begin{tabular}{@{}lccccc@{}}
\toprule
half-A predictor & $r$ & g-res.\ OOF & $R^2$ & RMSE & MAE \\
\midrule
$B_C$                        & 0.844 & 0.613 & 0.713 & 1.86 & 1.44 \\
$\Phi(\hat\bth)$             & 0.884 & 0.635 & 0.781 & 1.62 & 1.22 \\
measured FKGL$_A$            & 0.943 & 0.838 & 0.889 & 1.15 & 0.81 \\
genre                        & 0.807 & --- & 0.651 & 2.05 & 1.57 \\
genre\,+\,$B_C$              & 0.886 & 0.607 & 0.785 & 1.60 & 1.20 \\
genre\,+\,$B_C$\,+\,$\Phi$   & 0.900 & 0.661 & 0.810 & 1.51 & 1.12 \\
\bottomrule
\end{tabular}
\caption{BNC baseline audit (LDA $K{=}100$, seed 0), split-half protocol: cross-fitted
out-of-fold linear predictions of half-B FKGL from half-A features, as in
Table~\ref{tab:dr2}; all 3{,}021 documents (genres with $n<20$ pooled into an ``other''
dummy). Paired document bootstrap: $\Delta r(\Phi-B_C)=0.039$
$[0.031,0.048]$; $\Delta R^2$ of adding $\Phi$ to genre$+B_C$: $0.024$ $[0.018,0.031]$.
The whole-document analysis uses the \emph{full-input} prediction (as in the
whole-document columns of Table~\ref{tab:main}) and reaches $0.927$ ($\Phi$), $0.915$
(genre$+B_C$) and $0.938$ (genre$+B_C+\Phi$), with $\Delta R^2=0.041$ $[0.035,0.048]$.
Rows are per-fold calibrated, so residual correlations differ slightly from the
raw-prediction values of Sec.~\ref{sec:exp-genre} ($0.635$ here vs.\ $0.638$ there). The
genre-only residual entry is omitted: residualising the genre prediction by the same
grouping variable leaves only leave-fold-out estimation noise and is not interpretable.
\emph{Fit sensitivity} (same folds; the headline fit, $K{=}100$ seed 0, was fixed in
advance): split-half $\Delta R^2$ across the five $K{=}100$ fits (seeds 0--4) is $0.024$
$[0.018,0.031]$, $0.021$ $[0.015,0.027]$, $-0.007$ $[-0.014,-0.001]$, $0.024$
$[0.017,0.030]$ and $0.013$ $[0.007,0.020]$ (median $0.021$; the $K{=}50$ fit: $0.019$
$[0.014,0.025]$); whole-document $\Delta R^2$: $0.041$, $0.040$, $0.074$, $0.046$,
$0.061$ ($K{=}50$: $0.038$), positive in all six fits. Variational LDA converges to
restart-dependent local optima; the reversing fit allocates most weight to the
boundary coordinate (highest unmasked accuracy, $r=0.949$; lowest content-masked,
$0.860$), which masking exposes (Sec.~\ref{sec:whattheta}).}
\label{tab:bnc}
\end{table}

\paragraph{Word-only representation with a supervised rate probe.}
To check that nothing above is specific to placing $\eos$ in the vocabulary, we train LDA
on the word-only vocabulary and read the rates off a \emph{supervised probe}: a ridge
regression $\bth\mapsto(e,\sigma)$ fitted on the outer training folds only, with $\alpha$
chosen by 5-fold internal cross-validation over 13 log-spaced values in
$[10^{-3},10^{3}]$, features standardised with training-fold statistics, and $\hat e$
clipped to $[10^{-4},1-10^{-4}]$. On the content-masked \emph{whole-document} task this
reaches $r=0.845$ (genre-residual $0.607$) against the boundary-augmented model's $0.848$
--- the comparison quoted in the body, with input, protocol and document length identical.
Under the \emph{split} protocol the probe is sensitive to how its supervision is supplied,
because $\hat\bth$'s posterior concentration depends on document length: a probe fitted on
half-length inputs against those halves' own rates transfers to unseen documents at
$r=0.694$, one fitted against the training documents' \emph{other} halves reaches $0.801$
--- but the latter learns the half-to-half shift directly and is therefore not comparable
with the plug-in decoder --- and a probe fitted on whole documents and applied to halves
collapses ($0.183$). We therefore rest the word-only claim on the whole-document
comparison. The BNC replicates it: the whole-document probe reaches $0.911$ against the
boundary-augmented $0.908$ (matched-supervision split: $0.879$ vs $0.884$). In all
variants the probe, unlike the decoder, uses the training documents'
measured rates as supervision.
\FloatBarrier

\section{Robustness}
\label{app:robust}

\paragraph{Random seeds.} Refitting on the whole corpus with seeds $0,1,2$ gives
whole-corpus $r$ of 0.870/0.869/0.884 (LDA), 0.814/0.815/0.814 (ProdLDA) and 0.811/0.846/0.799 (ETM). The overall correlation is stable; per-topic quantities are much
less so. The number of LDA topics with $q_k\ge0.01$ is $13/19/20$ across the three seeds, and
the range of $\Phi(\mathbf{e}_k)$ over those topics moves from $[2.80,20.12]$ to
$[4.66,21.45]$ to $[2.48,16.10]$. What does stay stable is the range over the \emph{observed}
posterior support: the 1st--99th percentile of $\Phi(\hat\bth_d)$ is $[4.47,14.07]$,
$[4.66,14.18]$ and $[4.04,14.46]$. Table~\ref{tab:topics} should accordingly be read as an
illustration of the mechanism, not as a measurement of particular topics' grade levels.

\paragraph{Formula placeholders.} Brown renders mathematical formulae as placeholder tokens
(\texttt{**zg} and four others), which account for $0.006\%$ of tokens. Deleting all five
types leaves the corpus FKGL unchanged to two decimals ($10.27\pm3.22$), gives a
whole-corpus LDA correlation of $r=0.871$ (against $0.870$ with them),
and shrinks the well-formed per-topic $\Phi$ range only from $[2.80,20.12]$ to
$[2.80,19.31]$. The long sentences of the highest-$\Phi$ topic are therefore a property of
mathematical prose under this tokenisation, not of the placeholder tokens.

\paragraph{Leave-one-genre-out.} Fitting the topic model on one of Brown's two super-groups
and evaluating on the other is a hard transfer test, since $(\bq,\bs)$ must be estimated
from documents of a different kind. LDA trained on informative prose reaches $r=0.642$ on imaginative
prose ($n=126$), and trained on imaginative prose reaches $r=0.425$ on informative
prose ($n=374$); ProdLDA gives $0.429$ and $0.612$. The
mapping from topic vectors to grade levels therefore transfers across the genre divide, but
with a substantial loss relative to the in-domain numbers of Table~\ref{tab:main}.
\FloatBarrier

\section{Genre analysis}
\label{app:genre}

\begin{table}[h]
\centering\small
\setlength{\tabcolsep}{4pt}
\begin{tabular}{@{}lccccc@{}}
\toprule
& all & within & partial & inform. & imagin. \\
\midrule
LDA & 0.861 & 0.651 & 0.656 & 0.682 & 0.757 \\
ProdLDA & 0.809 & 0.548 & 0.523 & 0.528 & 0.727 \\
ETM & 0.805 & 0.500 & 0.498 & 0.507 & 0.721 \\
\bottomrule
\end{tabular}
\caption{Genre confound. \emph{within}: Fisher-$z$, size-weighted mean (weights $n_c-3$) of
the Pearson $r$ computed separately inside
each of the 15 Brown categories. \emph{partial}: $r$ after removing category means from
both the prediction and the target. \emph{inform.}/\emph{imagin.}: $r$ within Brown's two
super-groups. All predictions are out of fold; \emph{partial} uses whole-sample category
means.}
\label{tab:genre}
\end{table}
\FloatBarrier

\section{Sensitivity to the number of topics}
\label{app:k}

\begin{table}[h]
\centering\small
\setlength{\tabcolsep}{5pt}
\begin{tabular}{@{}lcccc@{}}
\toprule
$K$ & 10 & 25 & 50 & 100 \\
\midrule
LDA      & 0.837 & 0.868 & 0.870 & 0.879 \\
ProdLDA  & 0.806 & 0.809 & 0.814 & 0.814 \\
ETM      & 0.790 & 0.814 & 0.811 & 0.827 \\
\midrule
\multicolumn{5}{@{}l}{\emph{Spearman }$\rho$} \\
LDA      & 0.843 & 0.868 & 0.861 & 0.873 \\
ProdLDA  & 0.786 & 0.781 & 0.781 & 0.782 \\
ETM      & 0.712 & 0.794 & 0.774 & 0.777 \\
\bottomrule
\end{tabular}
\caption{Pearson $r$ (top) and Spearman $\rho$ (bottom) between $\Phi(\hat{\bth})$ and
measured FKGL as a function of the number of topics. Unlike the main results, these use the
in-sample protocol (model fitted on all 500 documents), so they are not comparable in level
to Table~\ref{tab:main}; they are reported only to show that the ordering and magnitude are
not artefacts of $K=50$.}
\label{tab:k}
\end{table}
\FloatBarrier

\section{Agreement as a function of segment length}
\label{app:length}

\begin{figure*}[t]
\centering
\includegraphics[width=\textwidth]{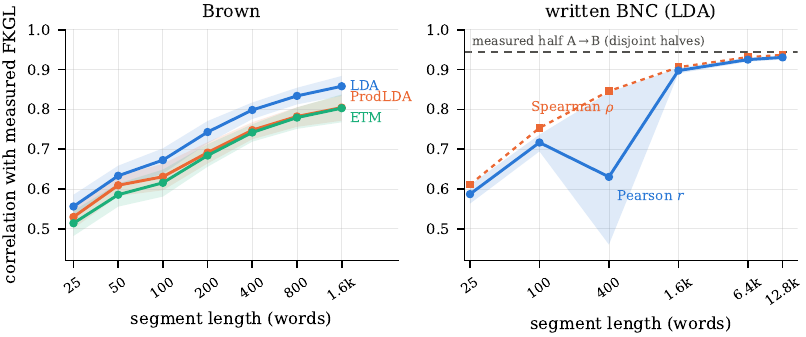}
\caption{Out-of-fold agreement as a function of segment length (whole-segment, full-input
evaluation; consecutive sentences from held-out documents; bands are segment-level
bootstrap 95\% CIs, optimistic because segments share documents). Left: Brown. Right: the
BNC continues the curve into the long-text regime; the dashed line is the
\emph{disjoint-halves} correlation of measured FKGL, a different protocol, shown for
orientation only. The Pearson dip at 400 words has a wide CI while Spearman $\rho$ is monotone,
indicating a small number of extreme segments; segment cohorts differ across lengths, so
the curve is descriptive.}
\label{fig:length}
\end{figure*}

Fig.~\ref{fig:length} plots the length curves discussed in Sec.~\ref{sec:exp-length}.

\section{Responsible Research Checklist Details}
\label{app:checklist}

\subsection{Scientific artifacts and release}
\label{app:checklist-artifacts}

Sec.~\ref{sec:setup} and App.~\ref{app:hp} identify and cite the Brown corpus, the
written BNC, NLTK, LDA, ProdLDA, ETM, scikit-learn and gensim. Syllable counts use
CMUdict version 0.7a \citep{cmudict}, distributed through NLTK's \texttt{cmudict}
corpus (123{,}455 word types), with a vowel-group fallback for uncovered word types.

The BNC licence does not permit redistribution of corpus text, and we redistribute none.
The released artifacts will contain only code, document identifiers and derived
statistics needed to reproduce the reported results from a licensed copy of the corpus.
The code and the derived result files will be made publicly available upon acceptance
under an open-source licence. The release will contain no person-level metadata, text
excerpts or free-text fields copied from either corpus.

\subsection{Model size and computational budget}
\label{app:checklist-compute}

All experiments ran on CPUs. The Brown experiments ran on a CPU-only x86-64 virtual
machine and completed in under two hours; each BNC fit ran on an AMD Ryzen Threadripper
PRO 7995WX (96 cores, 192 threads; 503~GiB RAM) and completed in under five hours on
48--50 workers (App.~\ref{app:hp}). Across the final models, the fit-sensitivity analyses, the topic-count
sensitivity analyses and the simulations, the total computational budget was
approximately 1{,}200 CPU-core-hours (wall-clock times multiplied by worker counts).
The fitted topic--word matrices have $698{,}800$ parameters (Brown LDA, $K{=}50$),
$7{,}500{,}200$ (BNC LDA, $K{=}100$) and $3{,}750{,}100$ (BNC LDA, $K{=}50$); on Brown,
ProdLDA has $6{,}504{,}076$ and ETM $7{,}586{,}628$ parameters including their amortised
inference networks (batch-normalisation layers are non-affine and contribute no learned
parameters).

\subsection{Software and package parameters}
\label{app:checklist-software}

The experiments used Python 3.10.9, NLTK 3.10.0, CMUdict 0.7a, scikit-learn 1.7.2,
NumPy 2.2.6 \citep{harris2020array}, SciPy 1.15.3 \citep{2020SciPy-NMeth} and joblib
1.5.3 \citep{joblib2025}; the neural topic models used PyTorch 2.13.0
\citep{NEURIPS2019_bdbca288}, and the ETM skip-gram initialisation used gensim 4.4.0.
The official project pages are NumPy (\url{https://numpy.org/}), SciPy
(\url{https://scipy.org/}), joblib (\url{https://joblib.readthedocs.io/}) and PyTorch
(\url{https://pytorch.org/}). Sec.~\ref{sec:setup} and App.~\ref{app:hp} report the
preprocessing and model parameters; a complete dependency list will be released with
the code.

\subsection{Use of AI assistants}
\label{app:checklist-ai}

AI assistants were used to review manuscript wording and mathematical exposition and to
assist with bibliographic verification and preparation of submission metadata and
checklist responses. Responsibility for the final claims, calculations, citations,
experimental results and wording remains with the authors.

\end{document}